\documentclass[runningheads,a4paper]{llncs}

\usepackage{amsmath}            
\usepackage{amssymb}            
\usepackage{array}              
\usepackage{multirow}
\usepackage{algorithm}          
\usepackage{algorithmic}        
\usepackage{booktabs}           

\usepackage{multicol}           
\usepackage[utf8]{inputenc}     
\usepackage[table]{xcolor}      
\usepackage{graphicx}

\makeatletter
\renewcommand\section{\@startsection{section}{1}{\z@}%
                       {-8\p@ \@plus -4\p@ \@minus -4\p@}%
                       {6\p@ \@plus 4\p@ \@minus 4\p@}%
                       {\normalfont\large\bfseries\boldmath
                        \rightskip=\z@ \@plus 8em\pretolerance=10000 }}
\renewcommand\subsection{\@startsection{subsection}{2}{\z@}%
                       {-8\p@ \@plus -4\p@ \@minus -4\p@}%
                       {6\p@ \@plus 4\p@ \@minus 4\p@}%
                       {\normalfont\normalsize\bfseries\boldmath
                        \rightskip=\z@ \@plus 8em\pretolerance=10000 }}
\renewcommand\subsubsection{\@startsection{subsubsection}{3}{\z@}%
                       {-4\p@ \@plus -4\p@ \@minus -4\p@}%
                       {-1.5em \@plus -0.22em \@minus -0.1em}%
                       {\normalfont\normalsize\bfseries\boldmath}}
\makeatother

\newcommand{\ReLU}{\text{ReLU}}
\newcommand{\tightsubsection}[1]{\par\medskip\noindent\textbf{#1~~}}

\usepackage{url}
\urldef{\mailsa}\path|{mohammad.havaei, nicolas.guizard, 
nicolas.chapados}@imagia.com|  
\newcommand{\keywords}[1]{\par\addvspace\baselineskip
\noindent\keywordname\enspace\ignorespaces#1}

\begin{document}

\mainmatter  

\title{HeMIS:\\Hetero-Modal Image Segmentation \thanks{Accepted as oral presentation at MICCAI 2016} \\}

\titlerunning{Hetero-Modal Image Segmentation}

%
%

\author{Mohammad Havaei \inst{1}\inst{2}\and Nicolas Guizard\inst{1}\and Nicolas Chapados\inst{1}\and Yoshua~ Bengio\inst{3}}

%
\authorrunning{Havaei, Guizard, Chapados, Bengio}

\institute{Imagia Inc., Montreal, Qc, Canada\\ 
\and Universit\'e de Sherbrooke, Qc, Canada \\
\and Universit\'e de Montr\'eal, Montr\'eal, Canada \\ 
\mailsa\\
\url{seyed.mohammad.havaei@usherbrooke.ca} \\
\url{yoshua.bengio@umontreal.ca}\\
\url{http://www.imagia.com}}

%
%

\toctitle{Lecture Notes in Computer Science}
\tocauthor{Authors' Instructions}
\maketitle

\begin{abstract}
We introduce a deep learning image segmentation framework that is extremely
robust to missing imaging modalities. Instead of attempting to impute or
synthesize missing data, the proposed approach learns, for each modality,
an embedding of the input image into a single latent vector space for which
arithmetic operations (such as taking the mean) are well defined. Points in
that space, which are averaged over modalities available at inference time,
can then be further processed to yield the desired segmentation. As such,
any combinatorial subset of available modalities can be provided as input,
without having to learn a combinatorial number of imputation
models. Evaluated on two neurological MRI datasets (brain tumors and MS lesions), the approach yields state-of-the-art segmentation results
when provided with all modalities; moreover, its performance degrades
remarkably gracefully when modalities are removed, significantly more so
than alternative mean-filling or other synthesis approaches.

\end{abstract}

\section{Introduction}
\label{sec:intro}

In medical image analysis, image segmentation is an important task and is
primordial to visualize and quantify the severity of the pathology in
clinical practice. Multi-modality imaging provides complementary
information to discriminate specific tissues, anatomies and
pathologies. However, manual segmentation is long, painstaking and subject
to human variability. In the last decades, numerous automatic approaches
have been developed to speed up medical image segmentation. These methods
can be grouped into two categories: The first, \emph{multi-atlas approaches}
estimate on-line intensity similarities between the
subject being segmented and multi-atlases (images with expert
labels). These techniques have shown excellent results in
structural segmentation when using non-linear registration
\cite{iglesias2015multi}; when combined with non-local approaches they have
proven effective in segmenting diffuse and sparse pathologies (ie. multiple
sclerosis (MS) lesions \cite{guizard2015rotation}) as well as more complex
multi-label pathology (ie. Glioblastoma \cite{cordier2016Patch}).
In contrast, \emph{model-based approaches} are
typically trained offline to identify a discriminative model of image
intensity features. These features can be predefined by the user
(e.g. with random forests \cite{geremia2013spatiallyRF}) or 
extracted and learned hierarchically directly from the images
\cite{Brosch2015MSencoder}. 

Both strategies are typically optimized for a specific set of multi-modal
images and usually require these modalities to be available. In clinical
settings, image acquisition and patient artifacts, among other hurdles,
make it difficult to fully exploit all the modalities; as such, it is
common to have one or more modalities to be missing for a given
instance. This problem is not new, and the subject of missing data analysis
has spawned an immense literature in statistics
(e.g. \cite{van2012flexible}). In medical imaging, a number of approaches
have been proposed, some of which require to re-train a specific model with
the missing modalities or to synthesize them
\cite{hofmann2008MRIPETsynthesis}. Synthesis can improve multi-modal
classification by adding information of the missing modalities in the
context of a simple classifier such as random forests
\cite{Tulder2015MRISynthesis}. Approaches to imitate with fewer features a
classifier trained with a complete set of features have also been proposed
\cite{hor2015scandent}.  Nevertheless, it should stand to reason that a
more complex model should be capable of extracting relevant features from
just the available modalities without relying on artificial intermediate
steps such as imputation or synthesis.

This paper proposes a deep learning framework (HeMIS) that can segment
medical images from incomplete multi-modal datasets. Deep learning
\cite{Goodfellow-et-al-2016-Book} has shown an increasing popularity in
medical image processing for segmenting but also to synthesize missing
modalities \cite{Tulder2015MRISynthesis}. Here, the proposed approach learns, separately for each 
modality, an embedding of the input image into a
latent space. In this space, arithmetic operations (such
as computing first and second moments of a collection of vectors) 
are well defined and can be taken
over the different modalities available at inference time. These computed moments
can then be further processed to estimate the final
segmentation. This approach presents the advantage of being robust to any
combinatorial subset of available modalities provided as input, without the
need to learn a combinatorial number of imputation models. 
We start by describing the method (\S\ref{sec:method}), follow with a description
of the datasets (\S\ref{sec:data}) and experiments (\S\ref{sec:results}) and finally 
offer concluding remarks (\S\ref{sec:concl}).


\section{Method}
\label{sec:method}

\subsection{Hetero-Modal Image Segmentation}

Typical convolutional neural network (CNN) architectures take a multiplane
image as input and process it through a sequence of convolutional layers
(followed by nonlinearities such as $\ReLU(\cdot) \equiv \max(0,\cdot)$),
alternating with optional pooling layers, to yield a per-pixel or per-image
output \cite{Goodfellow-et-al-2016-Book}. In such networks every input
plane is assumed to be present within a given instance: since the very
first convolutional layer mixes input values coming from all planes, any
missing plane introduces a bias in the computation that the network is not
equipped to deal with.

\begin{figure}[t]
    \includegraphics[width=\textwidth,height=\textheight,keepaspectratio]{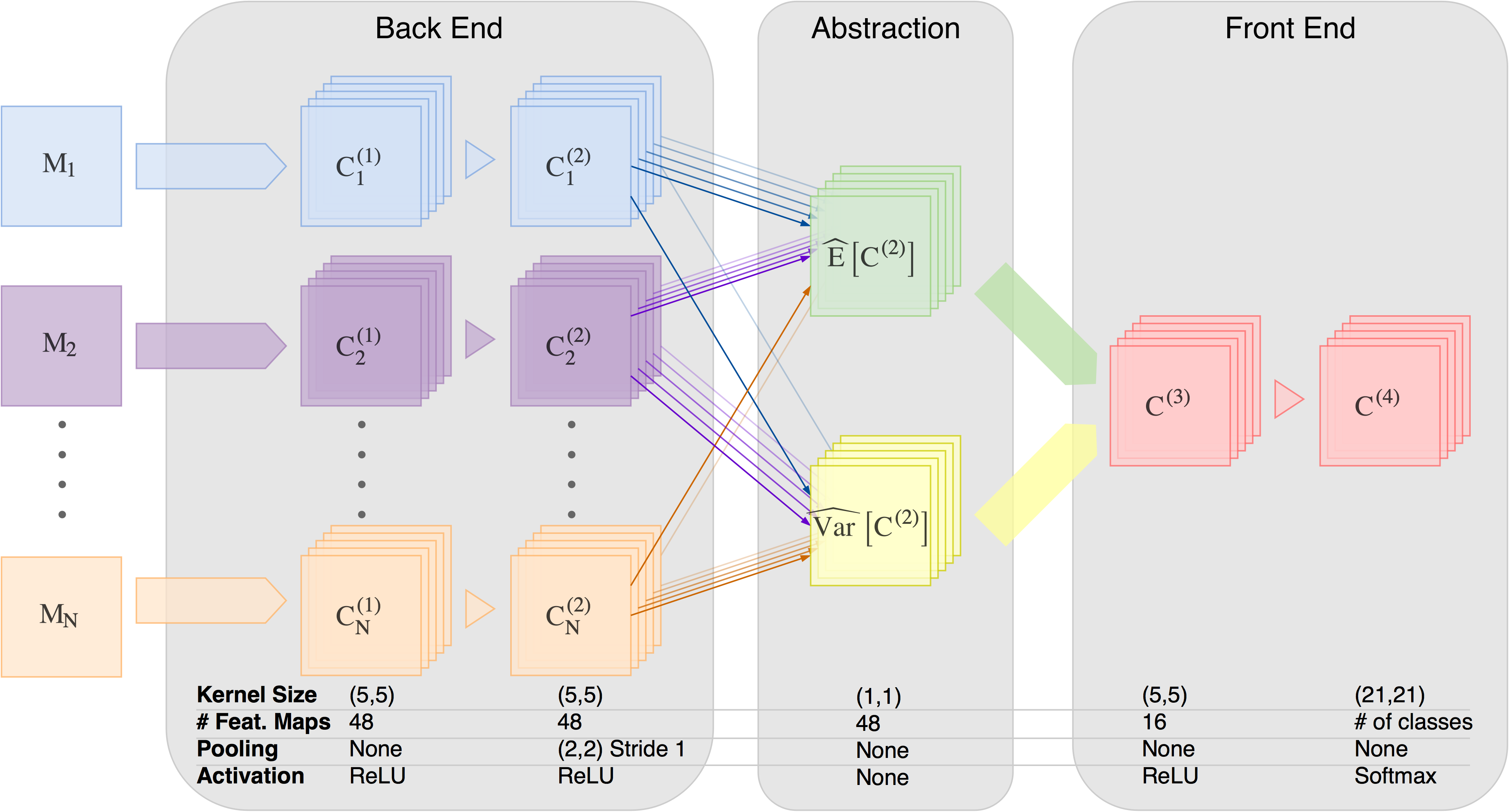}
    \caption{Illustration of the Hetero-Modal Image Segmentation
      architecture. Modalities available at inference time, $M_k$, are
      provided to independent modality-specific convolutional layers in the
      \textbf{back end}. Feature maps statistics (first \& second
      moments) are computed in the \textbf{abstraction layer}, which after
      concatenation are processed by further convolutional layers in the
      \textbf{front end}, yielding pixelwise classifications outputs.}
    \label{fig:hemis}
     
\end{figure}

We propose an approach wherein each modality is initially processed by its
own convolutional pipeline, independently of all others. After a few
independent stages, feature maps from all available modalities are
\emph{merged} by computing mapwise statistics such as the mean and the
variance, quantities whose expectation does not depend on the number of terms
(i.e. modalities) that are provided. After merging, the mean and variance
feature maps are concatenated and fed into a final set of convolutional
stages to obtain network output. This is illustrated in
Fig.~\ref{fig:hemis}.
In this procedure, each modality contributes a separate term to the
mean and variance; in contrast to a vanilla CNN architecture, a missing
modality does not throw this computation off: the mean and variance terms
simply become estimated with larger uncertainty. In seeking to be robust
to any subset of missing modalities, we call this approach
\emph{hetero-modal} rather than multi-modal, recognizing that in addition
to taking advantage of several modalities, it can take advantage of a
diverse, instance-varying, set of modalities. In particular, it does not
require that a ``least common denominator'' modality be
present for every instance, as sometimes needed by common imputation
methods.

Let $k \in \mathcal{K} \subseteq \{1, \ldots, N\}$ denote a modality within
the set of available modalities for a given instance, and $M_k$
represent the image of the $k$-th modality. For simplicity, in this work we
assume 2D data (e.g. a single slice of a tomographic image), but it
can be extended in an obvious way to full 3D sections. 
As shown on Fig.~\ref{fig:hemis}, HeMIS proceeds in three stages:

\par\smallskip\noindent\textit{1. Back End:} In our implementation, this consists of two
convolutional layers with \ReLU, the second followed with a $(2,2)$
max-pooling layer, denoted respectively $C_k^{(1)}$ and $C_k^{(2)}$. To
ensure that the output layer consists of the same number of pixels as the
input image, the convolutions are zero-padded and the stride for all
operations (including max-pooling) is 1. In particular, pooling with a stride
of 1 \emph{does not downsample}, but simply ``thickens'' the feature maps;
this is found to add some robustness to the results. The number of feature maps in each
layer is given in Fig.~\ref{fig:hemis}. Let $C_{k,\ell}^{(j)}$ be the the
$\ell$-th feature map of $C_k^{(j)}$.

\par\smallskip\noindent\textit{2. Abstraction Layer:} Modality fusion is computed here, as
first and second moments across available modalities in $C^{(2)}$,
separately for each feature map $\ell$,
\[
  \widehat{\mathrm{E}}_{\ell}\left[C^{(2)}\right] = 
    \frac{1}{|\mathcal{K}|} 
    \sum_{k \in \mathcal{K}} C_{k,\ell}^{(2)}
  \quad\text{and}\quad
  \widehat{\mathrm{Var}}_{\ell}\left[C^{(2)}\right] = 
    \frac{1}{|\mathcal{K}|-1} 
    \sum_{k \in \mathcal{K}} \left( C_{k,\ell}^{(2)} -
            \widehat{\mathrm{E}}_{\ell}\left[C^{(2)}\right] \right)^2\!\!,
\]
with $\widehat{\mathrm{Var}}_{\ell}[C^{(2)}]$ defined to be zero if
$|\mathcal{K}|=1$ (a single available modality).

\par\smallskip\noindent\textit{3. Front End:} Finally the front end combines the merged
modalities to produce the final model output. In our implementation, we
concatenate all $\widehat{\mathrm{E}}\left[C^{(2)}\right]$ and
$\widehat{\mathrm{Var}}\left[C^{(2)}\right]$ feature maps, pass them
through a convolutional layer $C^{(3)}$ with ReLU activation, to finish
with a final layer $C^{(4)}$ that has as many feature maps as there are
target segmentation classes. The pixelwise posterior class probabilities
are given by applying a softmax function across the $C^{(4)}$ feature
maps, and a full image segmentation is obtained by taking the pixelwise
most likely posterior class. No further postprocessing on the resulting
segment classes (such as smoothing) is done.

\tightsubsection{Pseudo-Curriculum Training Procedure}
To carry out segmentation efficiently, the model is trained fully 
convolutionnally to minimize a pixelwise class cross-entropy loss, in the 
spirit of \cite{long2015fully}.
It has long been known that noise injection during training is a powerful technique to make neural networks more robust, as shown among others with denoising autoencoders \cite{vincent2008extracting}, and dropout and related procedures \cite{srivastava2014dropout}.
Here, we make the HeMIS architecture robust to missing modalities by randomly dropping any number for a given training example.
Inspired by previous works on curriculum learning \cite{bengio2009curriculum}---where the model starts learning from easy scenarios  before turning to more difficult ones---we used a pseudo-curriculum learning scheme where after few warmup epochs where all modalities are shown to the model, we start randomly dropping modalities, ensuring a higher probability of dropping zero or one modality only.



\tightsubsection{Interpretation as an Embedding}
An embedding is a mapping from an arbitrary source space to a target
real-valued vector space of fixed dimensionality. In recent years,
embeddings have been shown to yield unexpectedly powerful representations
for a wide array of data types, including single words
\cite{mikolov2013distributed}, variable-length word sequences
and images \cite{icml2015_xuc15}, and more.

In the context of HeMIS, the back end can be interpreted as learning to
separately map each modality into an \emph{embedding common to all
modalities}, within which vector algebra operations carry well-defined 
semantics. As such, computing empirical moments to carry out modality fusion is
sensible. Since the model is trained entirely end-to-end with
backpropagation, the key aspect of the architecture is that this embedding
only needs be defined implicitly as that which minimizes the overall
training loss. Cross-modality interactions can be captured within specific
embedding dimensions, as long as there are a sufficient number of them
(i.e. enough feature maps within $C^{(2)}$), as they can be combined by 
$C^{(3)}$.

With this interpretation, the back end consists of a modular assembly of
operators, viewed as reusable building blocks that may or may not be needed
for a given instance, each computing the embedding from its own input modality.
These projections are
summarized in the abstraction layer (with a mean and variance, although
additional summary statistics are simple to entertain), and this summary
further processed in the front end to yield final model output.

%
%

%
%

\section{Data and Implementation details}
\label{sec:data}

We studied the HeMIS framework on two neurological pathologies: Multiple Sclerosis (MS) with the MS Grand Challenge (MSGC) and a large Relapsing Remitting MS (RRMS) cohort, as well as glioma with the Brain Tumor Segmentation (BRATS) dataset \cite{Menze2014Brats}. 

\par\noindent\textbf{MS}
\textit{MSGC:}
The MSGC dataset 
\cite{styner2008MSGC} provides 20 training MR cases with manual ground truth lesion 
segmentation and 23 testing cases from the Boston Children’s Hospital (CHB) and the University of North Carolina (UNC). We 
downloaded\footnote{\url{http://www.nitrc.org/projects/msseg/}} the co-registered T1W, T2W, FLAIR images for all 43 cases as well as the ground truth lesion mask images for the 20 training cases. While lesions masks for the 23 testing cases are not available for download, an automated system is available to evaluate the output of a given segmentation algorithm. 

\par\textit{RRMS:}
This dataset is obtained from a multi-site clinical study with 300 relapsing-remitting MS (RRMS) patients (mean age 37.5~yrs, SD 10.0~yrs). Each patient underwent an MRI that included sagittal T1W , T2W and T1 post-contrast (T1C) images. The MRI data were acquired on 1.5T scanners from different manufacturers (GE, Philips and Siemens). 

\par\noindent\textbf{BRATS}
The \textit{BRATS-2015} dataset contains 220 subjects with high grade and 54 subjects
with low grade tumors. Each subject contains four MR modalities (FLAIR, T1W,
T1C and T2) and comes with a voxel level segmentation ground truth of 5 labels: \textit{healthy}, \textit{necrosis}, \textit{edema}, \textit{non-enhancing tumor} and \textit{enhancing tumor}. 
As done in \cite{Menze2014Brats}, we transform each segmentation map into 3 binary maps which correspond to 3 tumor categories, namely; \textit{Complete} (which contains all tumor
classes), \textit{Core} (which contains all tumor subclasses except ``edema'') and \textit{Enhancing} (which includes the ``enhanced tumor'' subclass). For each binary map, the Dice Similarity Coefficient (DSC) is calculated \cite{Menze2014Brats}.

\textit{BRATS-2013} contains two test datasets; Challenge and Leaderboard. The Challenge dataset contains 10 subjects with high grade tumors while the Leaderboard dataset contains 15 subjects with high grade tumors and 10 subject with low grade tumors. There are no ground truth provided for these datasets and thus quantitative evaluation can be achieved via an online evaluation system~\cite{Menze2014Brats}. In our experiments we used Challenge and Leaderboard datasets to compare the HeMIS segmentation performance to the state-of-the-art, when trained on all modalities.

\tightsubsection{Pre-processing and implementation details}
Before being provided to the network, bias field correction \cite{sled1998n3} and intensity normalization with a zero mean, truncation of 0.001 quantile and unit variance is applied to the image intensity. The multi-modal images are co-registered to the T1W and interpolated to 1mm isotropic resolution.  

We used Keras library \cite{chollet2015Keras} for our implementation. To deal with class imbalance, we adopt the patch-wise training procedure mentioned in \cite{havaei2015}. We first train the model with a balanced dataset which allows learning features that are agnostic to the class distribution. In a second phase, we train only the final classification layer with a distribution close to the ground truth. This ensures that  we learn good features yet keep the correct class priors. The method was trained using an Nvidia TitanX GPU, with a stochastic gradient learning rate of $0.001$, decay constant of $0.0001$ and Nesterov momentum coefficient of $0.9$ \cite{sutskever2013}. For both BRATS-2015 and MS, we split the dataset into three separate subsets---train, valid and test---with ratios of 70\%, 10\% and 20\% respectively. To avoid over-fitting we used early stopping on the validation set.

\section{Experiments and Results}
\label{sec:results}

We first validate HeMIS performance against state-of-the-art segmentation methods on the two challenge datasets: MSGC and
BRATS. Since the test data and the ranking table for BRATS 2015
are not available, we submitted results to BRATS 2013 challenge and
leaderboard. These results are presented in Table \ref{tab:BraTS_submission}.\footnote{Note that the results mentioned in Table~\ref{tab:BraTS_submission} are from methods competing in the BRATS 2013 challenge for which a static table is provided at https://www.virtualskeleton.ch/BraTS/StaticResults2013. Since then, other methods have been added to the scoreboard but for which no reference is available.} As we observe, HeMIS outperforms Tustison et~al.~\cite{tustison2015optimal}, the winner of the BRATS 2013 challenge, on most tumor region categories.

The MSGC dataset illustrates a direct application of HeMIS flexibility as only three modalities (T1W, T2W and FLAIR) are provided for a small training set. Therefore, given the small number of subjects, we first trained HeMIS on RRMS dataset with 
four modalities
 and fine-tuned on MSGC. Our results were submitted to the MSGC website\footnote{http://www.ia.unc.edu/MSseg}, with a resuts summary appearing in Table~\ref{tab:fullresultsMSGC}. The MSGC segmentation results include three other supervised approaches; 
 when compared to them, HeMIS obtains highly competitive results with a combined score of 83.2\%, where 90.0\% would represent human performance given inter-rater variability.

\begin{table}[t!]
\begin{center}
    \caption {Comparison of HeMIS when trained on all modalities against BRATS-2013 Leaderboard and Challenge winners, in terms of Dice Similarity (scores from \cite{Menze2014Brats}). } \label{tab:BraTS_submission}
\newcolumntype{s}{>{\columncolor[gray]{0.9}} p{1.8cm}}%
\scalebox{1}{

\begin{tabular}{ s p{1.4cm} p{1.4cm} p{1.4cm} }
\hline
\rowcolor{lightgray} & \multicolumn{3}{c}{\textbf{Leaderboard}} 	\\
\hline
\rowcolor[gray]{0.9} Method & Complete & Core & Enhancing  \\
\hline
Tustison \cite{tustison2015optimal}	&79	           &65	        &53	\\  \hline               
Zhao \cite{Zhao2013semiautobrats}    	&79	           &59	        &47   \\ \hline
Meier \cite{Menze2014Brats}   	&72	&60	&53 \\ \hline
HeMIS       &\textbf{83}     &\textbf{67}   &\textbf{57}  \\
\hline
\end{tabular}
\begin{tabular}{| p{1.4cm} p{1.0cm} p{1.4cm} }
\hline
\rowcolor{lightgray} \multicolumn{3}{|c}{\textbf{Challenge}} 	\\
\hline
\rowcolor[gray]{0.9}Complete & Core & Enhancing \\\hline
87	            &\textbf{78}    &\textbf{74}	\\ \hline
84	            &70     	    &65	            \\ \hline
82	            &73	            &69	            \\ \hline
\textbf{88}     &75             &\textbf{74}    \\ \hline
\end{tabular}
}

\end{center}
\end{table}

\begin{figure}[t]
\centering
  \caption{MLP-imputed FLAIR for an MS patient. The figure shows from left to right the original modality and the predicted FLAIR given other modalities.}\label{fig:ms_imputed_modalities}
\includegraphics[width=0.85\textwidth,height=\textheight,keepaspectratio]{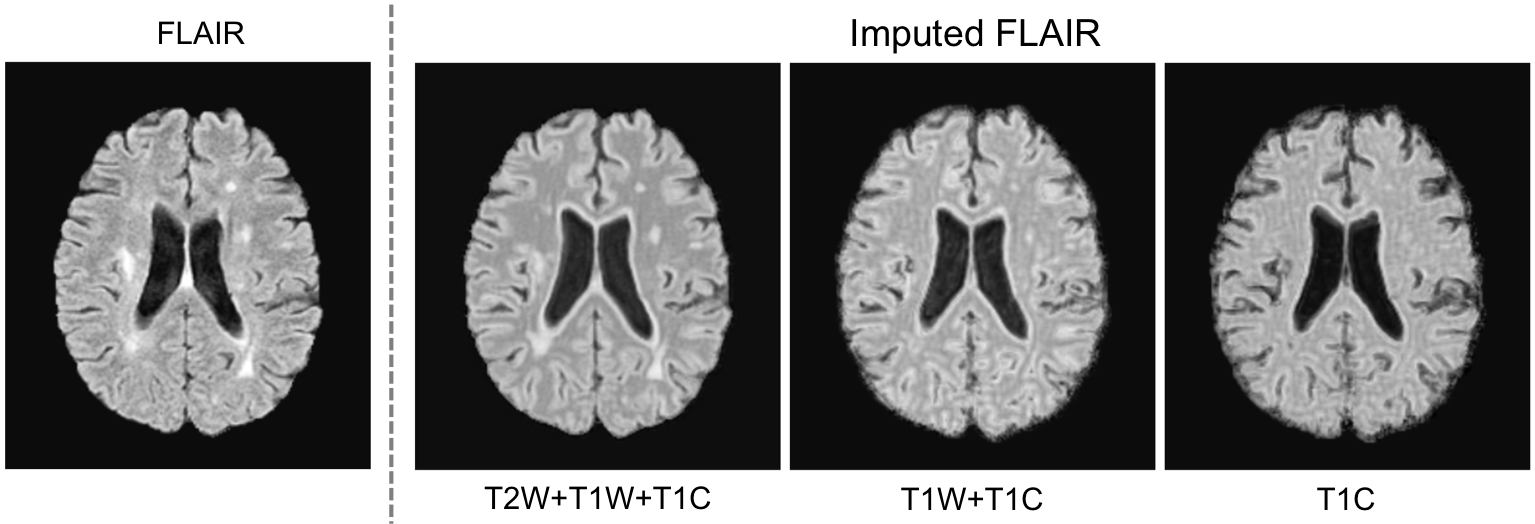}
\end{figure}

\begin{table}[t!]
\begin{center}
    \caption {Results of the full dataset training on the MSGC. For each rater (CHB and UNC), we provide the volume difference (VD), surface distance (SD), true positive rate (TPR), false positive rate (FPR) and the method's score as in \cite{styner2008MSGC}. } \label{tab:fullresultsMSGC}
\renewcommand{\arraystretch}{1.1}%
\newcolumntype{s}{p{1.4cm}}%
\scalebox{1}{
\begin{tabular}{ p{2.4cm} p{1.5cm}ssssl}
\hline
\rowcolor{lightgray}{} Method                                   &Rater  &VD (\%)&SD (mm)&TPR (\%)&FPR (\%)&Score\\
\hline
\cellcolor[gray]{0.9}
                                                                &CHB    &86.4   &8.4    &58.2   &70.6   &\multirow{2}{*}{80.0}\\
\multirow{-2}{*}{\cellcolor[gray]{0.9}Souplet et al. \cite{Souplet2008SupervisedMS}}  
                                                                &UNC    &57.9   &7.5    &49.1   &76.3   &\\
\hline
\cellcolor[gray]{0.9}
                                                                &CHB    &\textbf{52.4}   &\textbf{5.4}    &59.0    &71.5   &\multirow{2}{*}{82.1}\\
\multirow{-2}{*}{\cellcolor[gray]{0.9}Geremia et al. \cite{geremia2013spatiallyRF}}   
                                                                &UNC    &\textbf{45.0}   &\textbf{5.7}    &51.2    &76.7   &\\
\hline
\cellcolor[gray]{0.9}
                                                                &CHB    &63.5   &7.4    &47.1   &\textbf{52.7}   &\multirow{2}{*}{\textbf{84.0}}\\ 
\multirow{-2}{*}{\cellcolor[gray]{0.9}Brosch et al. \cite{Brosch2015MSencoder}}        
                                                                &UNC    &52.0   &6.4    &\textbf{56.0}   &\textbf{49.8}    &\\
                                                                
\hline
\cellcolor[gray]{0.9}
                                                                &CHB    &127.4  &7.5    &\textbf{66.1}   &55.3   &\multirow{2}{*}{83.2}\\
\multirow{-2}{*}{\cellcolor[gray]{0.9}HeMIS}                                          
                                                                &UNC    &68.2   &6.6    &52.3   &61.3   &\\\hline
\end{tabular}
}
\end{center}
\end{table}

The main advantage of HeMIS lies in its ability to deal with missing modalities,
specifically when different subjects are missing different modalities. 
To illustrate the model's flexibility in such circumstances, we compare HeMIS
performance to two common approaches to deal with random missing modalities. 
The first, \emph{mean-filling}, is to replace a missing
modality by the modality's mean value. In our case since all means are zero by construction, 
replacing a missing modality by zeros can be viewed as imputing with the mean. 
The second approach is to train a \emph{multi-layer perceptron} (MLP) to predict the expected value of specific missing modality given the available ones. Since neural networks are generally trained for a unique task, we need to train 28 different MLPs (one for each $\circ$ in Table~\ref{tab:big_table} for a given dataset) to account for different possibilities of missing modalities.
We used the same MLP architecture for all these models, which consists of 2 hidden layers with 100 hidden units each, trained to minimize the mean squared error. Fig.~\ref{fig:ms_imputed_modalities} shows an example of predicted modalities for an MS patient. 

\begin{table}[t]
\centering
\small
\caption {Dice similarity coefficient (DSC) results on the RRMS and BRATS test sets (\%) when modalities are dropped. The table shows the DSC for all possible configurations of MRI modalities being either absent~($\circ$) or present~($\bullet$), in order of FLAIR ($F$), T1W ($T_1$), T1C ($T_1c$), T2W ($T_2$). Results are reported for HeMIS, Mean (mean-filling) and the imputation MLP (MLP).}\label{tab:big_table} 
 
\newcolumntype{s}{p{1.0cm}}
\renewcommand{\arraystretch}{1.2}
\newcolumntype{L}{>{\columncolor[gray]{0.9}$}c<{$}}
\scalebox{0.78}{
\begin{tabular}{ LLLL|sss|sss|sss|sss}
\cline{5-16}
\rowcolor{lightgray}{} \cellcolor{white} & \cellcolor{white} & \cellcolor{white} & \cellcolor{white} & \multicolumn{3}{c|}{RRMS}& \multicolumn{9}{c}{BRATS} \\
\hline
\rowcolor[gray]{0.9} \multicolumn{4}{c|}{Modalities} & \multicolumn{3}{c|}{Lesion}& \multicolumn{3}{c|}{Complete} & \multicolumn{3}{c|}{Core}  &  \multicolumn{3}{c}{Enhancing}\\
\hline
\rowcolor{lightgray} 
\;F\;&\,T_1\,&T_1c&\,T_2\,&HeMIS  &Mean   &MLP    &HeMIS  &Mean   &MLP    &HeMIS  &Mean &MLP    &HeMIS  &Mean &MLP\\
\hline
\circ   & \circ   & \circ   & \bullet    &1.74   &2.66   &\textbf{12.77}  &58.48  &2.70   &\textbf{61.50}  &\textbf{40.18}  &4.00   &37.32  &\textbf{20.31}  &6.25   &18.62\\
\rowcolor[gray]{0.9}
\circ   & \circ   & \bullet & \circ      &2.67   &0.00   &\textbf{3.51}   &\textbf{33.46}  &23.11  &2.04   &\textbf{44.55}  &23.90  &17.70  &\textbf{49.93}  &30.02  &32.92\\
\circ   & \bullet & \circ   & \circ      &3.89   &0.00   &\textbf{6.64}   &\textbf{33.22}  &0.00   &2.07   &\textbf{17.42}  &0.00   &10.52  &4.67   &6.25   &\textbf{10.78}\\
\rowcolor[gray]{0.9}
\bullet & \circ   & \circ   & \circ      &34.48  &9.77   &\textbf{38.46}  &71.26  &\textbf{72.30}  &63.81  &\textbf{37.45}  &0.00   &34.26  &5.57   &6.25   &\textbf{15.90}\\
\circ   & \circ   & \bullet & \bullet    &\textbf{27.52}  &4.31   &25.83  &\textbf{67.59}  &35.01  &64.97  &\textbf{63.39}  &30.92  &49.38  &\textbf{65.38}  &39.00  &60.30\\
\rowcolor[gray]{0.9}
\circ   & \bullet & \bullet & \circ      &8.21   &0.00   &\textbf{8.26}   &\textbf{45.93}  &23.63  &1.99   &\textbf{55.06}  &41.89  &26.55  &\textbf{62.40}  &43.80  &40.93\\
\bullet & \bullet & \circ   & \circ      &38.81  &11.62  &\textbf{39.15}  &\textbf{80.28}  &75.58  &78.13  &\textbf{49.52}  &0.00   &48.97  &22.26  &6.25   &\textbf{25.18}\\
\rowcolor[gray]{0.9}
\circ   & \bullet & \circ   & \bullet    &\textbf{31.25}  &8.31   &29.39  &\textbf{69.56}  &1.77   &66.88  &\textbf{47.26}  &2.63   &43.66  &23.56  &6.25   &\textbf{26.37}\\
\bullet & \circ   & \circ   & \bullet    &\textbf{39.64}  &33.31  &38.55  &\textbf{82.1}   &81.01  &81.35  &\textbf{53.42}  &25.94  &52.41  &23.19  &6.25   &\textbf{25.01}\\
\rowcolor[gray]{0.9}
\bullet & \circ   & \bullet & \circ      &\textbf{41.38}  &6.42   &39.33  &79.8   &45.97  &\textbf{81.13}  &\textbf{66.12}  &29.85  &65.51  &\textbf{67.12}  &35.14  &66.19\\
\bullet & \bullet & \bullet & \circ      &\textbf{41.97}  &9.00   &40.63  &80.88  &81.57  &\textbf{82.19}  &69.26  &62.13  &\textbf{69.34}  &\textbf{71.30}  &67.13  &70.93\\
\rowcolor[gray]{0.9}
\bullet & \bullet & \circ   & \bullet    &\textbf{46.6}   &41.12  &41.83  &\textbf{83.87}  &77.84  &80.40  &\textbf{57.76}  &20.66  &53.46  &\textbf{28.46}  &6.25   &28.34\\
\bullet & \circ   & \bullet & \bullet    &\textbf{41.90}  &38.95  &41.47  &82.78  &64.19  &\textbf{83.37}  &\textbf{70.62}  &42.36  &70.45  &70.52  &49.62  &\textbf{70.56}\\
\rowcolor[gray]{0.9}
\circ   & \bullet & \bullet & \bullet    &\textbf{34.98} &5.78   &29.46   &\textbf{70.98}  &30.86  &67.85  &\textbf{66.60}  &45.79  &55.40  &\textbf{67.84}  &50.21  &64.81\\
\bullet & \bullet & \bullet & \bullet    &\textbf{48.66}  &43.48  &43.48  &\textbf{83.15}  &82.43  &82.43  &\textbf{72.5}   &71.46  &71.46 &\textbf{75.37}     &72.08  &72.08\\
\hline
\rowcolor[gray]{0.9}
\multicolumn{4}{c|}{\# Wins~/~15} & 
\multicolumn{1}{c}{\textbf{9}} & \multicolumn{1}{c}{0} & \multicolumn{1}{c|}{6} & 
\multicolumn{1}{c}{\textbf{10}} & \multicolumn{1}{c}{1} & \multicolumn{1}{c|}{4} & 
\multicolumn{1}{c}{\textbf{14}} & \multicolumn{1}{c}{0} & \multicolumn{1}{c|}{1} & 
\multicolumn{1}{c}{\textbf{9}} & \multicolumn{1}{c}{0} & \multicolumn{1}{c}{6} \\ \hline
\end{tabular}
}
 
\end{table}

Table~\ref{tab:big_table} shows the DSC for this experiment on the test set. 
On the BRATS dataset, for the Core category, HeMIS achieves the best segmentation in almost all cases (14 out of 15) 
and for the Complete and Enhancing categories it leads in most cases (10 and 9 cases out of 15 respectively). 
Also, the mean-filling approach hardly outperforms HeMIS or MLP-imputation. These results are consistent with the MS lesion segmentation dataset, where HeMIS outperforms other imputation approaches in 9 out of 15 cases. 
In scenarios where only one or two modalities are missing, while both HeMIS and MLP-imputation obtain good results, HeMIS outperforms the latter in most cases on both datasets. 
On BRATS, when missing 3 out of 4 modalities, HeMIS outperforms the MLP in a majority of cases. Moreover, whereas the HeMIS performance only gradually drops as additional modalities become missing, the performance drop for MLP-imputation and mean-filling is much more severe. On the RRMS cohort, the MLP-imputation appears to obtain slightly better segmentations when only one modality is available.

\begin{figure}[t]
    \centering    
    \includegraphics[width=0.85\textwidth,height=\textheight,keepaspectratio]{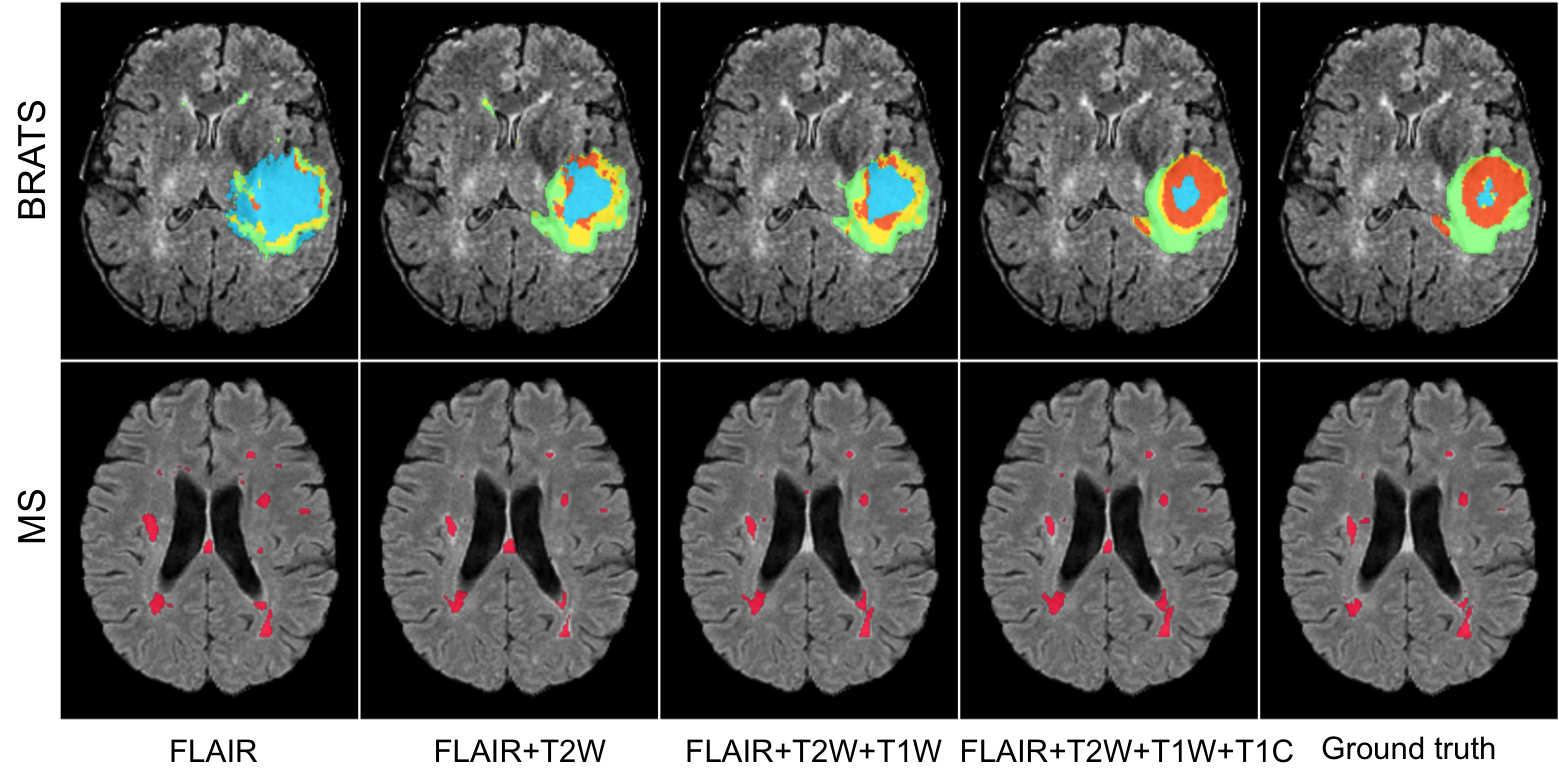}
     
    \caption{Example of HeMIS segmentation results on BRATS and MS subjects for different combinations of input modalities. For both cohorts, an axial FLAIR slice of a subject is overlaid with the results where for BRATS (first row) the segmentation colors describe necrosis (blue), non-enhancing (yellow), active core (orange) and edema (green). For the MS case, the lesions are highlighted in red. The columns present the results for different combinations of input modalities, with ground truth in the last column.}
     
\label{fig:seg_result}
\end{figure}

Although it is expected that tumor sub-label segmentations should be less accurate with fewer modalities, we should still hope for the model to report a sensible characterization of the tumor ``footprint''.
While MLP and mean-filling fail in this respect, HeMIS quite well achieves this goal by outperforming in almost all cases of the Complete and Core tumor categories. This can also be seen in Fig.~\ref{fig:seg_result} where we show how adding modalities to HeMIS improves its ability to achieve a more accurate segmentation. From Table~\ref{tab:big_table}, we can also infer that the FLAIR modality is the most relevant for identifying the Complete tumor while T1C is the most relevant for identifying Core and Enhancing tumor categories. On the RRMS dataset, HeMIS results are also seen to degrade slower than the other imputation approaches, preserving good segmentation when modalities go missing. Indeed, as seen in Fig.~\ref{fig:seg_result}, even though with FLAIR alone HeMIS already produces good segmentations, it is capable of further refining its results when adding  modalities, by removing false positives and improving outlines of the correctly identified lesions or tumor.

\section{Conclusion}
\label{sec:concl}

We have proposed a new fully automatic segmentation framework for heterogenous multi-modal MRI using a specialized convolutional deep neural network. The embedding of the multi-modal CNN back-end allows to train and segment datasets with missing modalities. We carried out an extensive validation on MS and glioma and achieved state-of-the art segmentation results on two challenging neurological pathology image processing tasks. Importantly, we contrasted the graceful performance degradation of the proposed approach as modalities go missing, compared with other popular imputation approaches, which it achieves without requiring training specific models for every potential missing modality combination. Future work should concentrate on extending the approach to broader modalities outside of MRI, such as CT, PET and ultrasound.

\bibliographystyle{splncs03}

\end{document}